\documentclass{article}
\usepackage[preprint]{neurips_ACA_2025}

\usepackage[utf8]{inputenc}
\usepackage[T1]{fontenc}
\usepackage{graphicx}
\usepackage{subcaption}
\usepackage{amsmath}
\usepackage{amsfonts}  
\usepackage{booktabs}
\usepackage{array}
\usepackage{tabularx}
\usepackage{microtype}
\usepackage{hyperref}
\usepackage[svgnames]{xcolor}
\usepackage{xstring}
\usepackage{xcolor}
\usepackage[normalem]{ulem}

\title{A Human Behavioral Baseline for Collective Governance in Software Projects}

\author{
  Mobina Noori \\
  Department of Computer Science\\
  University of California Davis  \\
  Davis, CA, USA \\
  \texttt{lianoori@ucdavis.edu} \\
  \And
  Mahasweta Chakraborti \\
  Department of Communication \\
  University of California, Davis  \\
  Davis, CA, USA \\
  \texttt{mchakraborti@ucdavis.edu} \\
  \And
  Amy X. Zhang \\
  Allen School of Computer Science \& Engineering \\
  University of Washington \\
  Seattle, WA, USA \\
  \texttt{axz@cs.uw.edu} \\
  \And
  Seth Frey \\
  Department of Communication \\
  University of California, Davis  \\
  Davis, CA, USA \\
  \texttt{sethfrey@ucdavis.edu} \\
}

\begin{document}
\maketitle

\begin{abstract}
% We treat AI systems as non-human stakeholders whose programmed objectives interact with human aims in shared workspaces. This view of algorithmically mediated collaboration motivates institutional oversight for both people and machines. To ground that oversight, we analyze version-controlled governance documents (\texttt{GOVERNANCE.md}) from open-source projects, comparing each repository’s earliest available text to its most recent. Our corpus precedes widespread adoption of AI management tools, providing a neutral baseline for future evaluation. By tracing how roles, responsibilities, and decision rights are written and revised, we offer an account of governance as a collectively negotiated artifact rather than a byproduct of individual behavior. We also contribute a replicable, text-based framework that can extend beyond markdown files to other governance records (e.g., instructions used to steer AI-assisted workflows). Together, these contributions establish a foundation for participatory AI research and supply benchmarks for assessing whether AI-mediated workflows concentrate authority or make participation more uneven relative to historical practice.
We study how open source communities describe participation and control through version controlled governance documents. Using a corpus of 710 projects with paired snapshots, we parse text into actors, rules, actions, and objects, then group them and measure change with entropy for evenness, richness for diversity, and Jensen Shannon divergence for drift. Projects define more roles and more actions over time, and these are distributed more evenly, while the composition of rules remains stable. These findings indicate that governance grows by expanding and balancing categories of participation without major shifts in prescriptive force. The analysis provides a reproducible baseline for evaluating whether future AI mediated workflows concentrate or redistribute authority.
\end{abstract}

\section{Introduction}

The use of artificial intelligence systems in software project management has become increasingly salient \citep{hashimzai2024integration}. In addition to assisting individual developers, they are coordinating core management functions, including drafting pull requests, triaging issues, proposing code reviews, and enforcing release gates. As these capabilities are embedded in team tooling, decision rights migrate from human maintainers toward sociotechnical pipelines. In these pipelines, algorithms and people jointly govern workflows \citep{xiao2024generative, wessel2022quality}. This shift raises questions about how authority is redistributed when algorithms mediate both individual contributions and collective coordination \citep{crawford2023ai}. Insights from recent work on algorithmic collective action indicate that when multiple groups interact with the same algorithmic system, their strategies can interfere in unexpected ways. A campaign that achieves near perfect success in isolation may see its efficacy drop sharply when a second group acts at the same time \citep{karan2025algorithmic}. We use \emph{algorithmic collective action} to denote coordinated behavior among participants when interactions are shaped by algorithmic systems within sociotechnical platforms. In software development, the growing reality of human and algorithmic co-production leads us to ask: how might AI systems embedded in team support tools reshape governance structures, stakeholder participation, and power relations on platforms, and what options exist for steering them toward the common good \citep{varanasi2023currently}?

The debate about how AI systems reshape governance has outpaced empirical evidence \citep{delgado2023participatory}. Scholars of AI governance and participatory design warn that algorithmic infrastructures can undermine stakeholder agency, reinforce hierarchies, or reallocate decision rights \citep{birhane2022power}. The recency of AI technologies make it currently challenging to gain substantial inferences on trends of human-AI interactions in software project management. In addition, most studies examine these emergent dynamics through case studies, audits, or simulations without a historical baseline against which to evaluate the change \citep{margetts2022rethinking}. As a result, claims about AI-induced shifts in governance remain speculative. Establishing a historical baseline for governance change is therefore the central objective of this paper. Before assessing whether AI systems redistribute power, narrow opportunities for participation, or enable more inclusive governance, it is necessary to understand how authority has evolved in primarily human-governed settings \citep{sharma2020artificial}. Our study addresses this need by constructing a large-scale longitudinal baseline of institutional change in open source project governance before the widespread adoption of AI-managed tools, providing a reference point for evaluating how future AI-mediated platforms may reshape participation and oversight. To our knowledge, this is the first large-scale longitudinal baseline of open-source governance prior to the widespread adoption of AI-managed tools.

Open source software communities have been extensively studied as exemplar instances of collective knowledge work~\cite{benkler2006wealth, heckman2007emergent, 10.1287/orsc.14.6.633.24866,10.5465/amj.2007.27169153, schweik2012internet, chakraborti2024we}. Importantly, they are a transparent testbed for studying governance \citep{o2005nonprofit}. Open source projects externalize governance in version-controlled files such as \texttt{GOVERNANCE.md}, which makes rules explicit, textual, and historically archived. Because governance edits are version-controlled and public, rule changes are observable at fine temporal resolution and comparable across time and projects. This is unique to OSS, which, unlike conventional organizational settings, supports systematic observation of how authority changes over time.

We treat AI systems as non-human stakeholders whose programmed objectives interact with human goals in shared workspaces. This framing aligns with a view of algorithmically mediated collaboration in which both human contributors and AI systems participate in shaping collective outcomes and therefore require institutional oversight. We lay a fundamental step in this important discourse, by analyzing version-controlled \texttt{GOVERNANCE.md} documents from open source software projects and contribute the following:

\begin{enumerate}

\item Our corpus captures several years of Open source software (OSS) projects before the widespread adoption of AI-managed project tools and management suites, offering a neutral reference point for future evaluations of AI-mediated governance.

\item  By tracing how roles, responsibilities, and decision rights evolve, we provide an account of governance as it is encoded and renegotiated collectively, rather than inferred only from individual behaviors or outcomes.

\item  We establish a text-based analytical framework that is replicable and easily extendable to governance records besides markdown files (e.g. prompts used to steer agentic workflows), and therefore can support future studies aimed at understanding software engineering team power structures under AI agent-human co-production.
\end{enumerate}

Together these contributions establish a foundation for participatory AI research. Understanding organic governance trajectories in open source communities can inform the design of participatory AI systems that allow collective human input in decision processes and provide benchmarks for assessing whether AI infrastructure serves the common good. This baseline enables falsifiable pre/post evaluations of AI-mediated workflows, including whether authority becomes more concentrated or participation more uneven when assistants are introduced.

Building on this baseline, we frame our analysis around three research questions. First, how do these communities distribute authority over time, and what does that suggest for steering AI systems toward the common good? Second, how are norms, responsibilities, and decision rights encoded in open source governance over time and what parallels exist for encoding values into AI systems? Third, can open source governance evolution serve as a model for participatory AI design in which users collectively influence system behavior?

These questions move from describing historical change in open source governance to identifying patterns that matter for the future of AI systems. By tracing how authority shifts, how rules harden or soften, and which governance elements remain stable versus contested, we offer an empirical foundation for examining how AI-managed infrastructures may redistribute power, reshape participation, and influence the prospects for collective oversight.

The rest of the paper is organized as follows. The description of the data and methods is summarized in Section 2. Section 3 represents the main results of the study. Then, Section 4 provides interpretation. Finally, Section 5 describes the conclusion, limitation, and future work.

\section{Methods}
We describe the corpus, selection criteria, preprocessing, institutional parsing, and analysis that convert governance prose into comparable structures. The aim is a simple pipeline that others can rerun on future AI-managed cohorts.

\textbf{Data and coverage.} GitHub is the most widely used hosting platform for open-source development, built on the distributed version-control system Git. It provides an infrastructure for collaboration, coordination, and community visibility as well as storing code. Governance is a persistent concern in this context: projects must determine how authority should be allocated, how contributor rights should be granted or lost, and how conflicts should be resolved. Many communities address these governance challenges through informal norms, foundation-level oversight, and increasingly, explicit written constitutions. A notable development has been the emergence of \texttt{GOVERNANCE.md} as a de facto standard for codifying project rules, alongside related artifacts such as \texttt{CONTRIBUTING.md}, codes of conduct, and maintainership guides. These files articulate roles, permissions, obligations, and protected resources, making governance unusually transparent and traceable.

Starting with a seeded collection and filename patterns, we analyzed 710 repositories with at least one governance file at the repository root. The corpus spans 2013–2022, with governance commits recorded through June 2022 (earliest: 2013-05-09; latest: 2022-05-19). File coverage is dominated by \texttt{GOVERNANCE.md}, which appears in 673 out of 710 projects (94.8\%), alongside 37 filename variants. The latest governance file is present for all projects, and Markdown structure is detectable in 498 repositories (70.1\%), with a median of 5 sections (range: 2 to 25) \cite{10174145}.

Across the corpus we record 3,889 governance commits corresponding to repository by commit pairs and 2,890 unique commit OIDs, covering 107,869 line level edits with 82,076 additions and 25,793 deletions.

\textbf{Paired subset and pairing rules.} The pipeline produced net governance changes (earliest and latest snapshots) for the 637 repositories over an observation period from 2014-03-26 to 2022-05-18. \texttt{GOVERNANCE.md} file names were dominant, being 601 of 637 (or 94.3\%). Inclusion requires at least two recoverable governance snapshots per project; we label the earliest valid snapshot as \emph{initial} and the most recent as \emph{latest}. We require across day change with the two snapshots fall on different calendar days, which in practice implies at least two distinct \texttt{GOVERNANCE.md} commits. For each repository with a governance file, we traversed the Git history to recover the earliest valid version of that file and paired it with the most recent version. Projects with only one usable snapshot are excluded from longitudinal analysis but retained for descriptive statistics. For the 637 paired repositories, the gap from earliest to latest commit has median 0 days with interquartile range 247, minimum 0, and maximum 2616; we refer to 0 day gaps as within day revisions. The across day change subset comprises 279 of 637 which equals 43.8 percent. Where multiple governance files existed, we would create a composite governance view by concatenating in a deterministic order and removing repeated boilerplate; in this paired cohort, one governance file per repository sufficed.

\textbf{Normalization and alignment.} We preprocess governance documents by removing badges and images, converting tables to lists, normalizing headings, and stripping markup. Text is segmented into short paragraph blocks and sentences using a splitter tuned for Markdown lists. To reduce pronoun ambiguity, we apply coreference resolution while maintaining a reversible mapping to original offsets \citep{lee-etal-2018-higher} \citep{jurafskyspeech}. Where headings are detectable, we record section counts across snapshots to capture the degree of governance structuring.

\textbf{Institutional parsing.} 
The governance structure of each GitHub project in our corpus was extracted from its \texttt{GOVERNANCE.md} constitution using the Institutional Grammar (IG) framework \citep{ostrom2009understanding} (Crawford \& Ostrom, 1995; Ostrom, 2006). This framework maps the syntactic elements of policy texts to institutional primitives, first decomposing paragraphs and multi-phrase sentences into simple "institutional statements". Under the institutional grammar, an institution is treated as a bag of institutional statements. Recent NLP methods have made their automated extraction from policy text feasible~\cite{10.1111/padm.12711,chakraborti2024nlp4gov}.
Governance documents were parsed into institutional statements consisting of four linked components. An institutional statement has a \emph{Role} (known in IG as 'Attribute') when its grammatical subject is a kind of agent. Roles account for the types of actor or position recognized by the institution (e.g., "project lead," "contributor"). An \emph{Action} (the `Aim' in IG; syntactically the verb) identifies activities recognized by the institution as requiring governance (e.g., "commit," "assign," "review"). A \emph{Deontic} captures the prescriptive force of the institutional statement, expressed through modal verbs such as "may," "should," or "must," which indicate whether an action is permitted, recommended, or required. Deontics can also be enabling ("can") or restricting ("cannot"). An \emph{Object} represents the grammatical object of the rule, whether another actor or a resource that is subject to the action enacted by the statement's role. For example, in the sentence "The technical committee must ratify the development roadmap", the Role is "technical committee," the Action is "ratify," the Object is "roadmap," and the Deontic is "must," which renders the statement obligatory. 

We extracted these components with the NLP4Gov toolkit \citep{chakraborti2024nlp4gov} \citep{chakraborti2024we}, which combines dependency parsing with semantic role labeling to parse each unitary institutional statement into its IG components. The parser emits tuples with and anchors spans and positions, enabling traceability back to the original text. Modal verbs such as \emph{may}, \emph{can}, \emph{should}, \emph{must}, and \emph{will} were canonicalized into a closed set of deontic types, while role names such as \emph{maintainer}, \emph{committer}, \emph{reviewer}, and \emph{release manager} were normalized into a controlled vocabulary manually. We further manually categorized Actions into a version of the Typology of Rules adapted from the institutional analysis literature~\cite{ostrom2009understanding, weible2012understanding, weible2017policy, weible2018understanding}. To test the reliability of these qualitative steps of the analysis, two authors coded the same sample of 50 Actions, which demanded the most manual categorization among the four types of institutional features. Over this sample they demonstrated a percent agreement of $82\%$ and a Cohen's $\kappa=0.92$, over 9 labels, including a null label, strong evidence for the intersubjective validity of the chosen typology.

\textbf{Embedding, clustering, and metrics.}
Each canonical tuple is rendered as a short governance statement.
We encode each governance statement with a Sentence-BERT encoder \citep{reimers2019sentence} and apply \textsc{BERTopic} \citep{grootendorst2022bertopic} to derive semantic clusters per repository at two snapshots (initial, latest).
\textsc{BERTopic} operates in the embedding space to form topic–like groups and uses class–based TF–IDF to label them. To ensure even clustering across all the projects' corpus, we use the library’s standard hyperparameters without custom tuning.
For structure, we report (i) richness $K$ as the number of distinct clusters per repository and (ii) normalized Shannon entropy $H$ (bits, base~2) over cluster proportions, with longitudinal change $\Delta H = H_{\text{latest}} - H_{\text{initial}}$.
For drift, we compute Jensen–Shannon divergence (JSD, bits) between the aligned initial and latest cluster distributions for each repository.
All repository–level estimates are aggregated with equal–weight bootstrap confidence intervals by resampling repositories with replacement.

% \textbf{Quality control, robustness, and reproducibility.}  

\textbf{Analysis and inference.}
Using the paired across day subset defined above, we compute, for each repository $r$ and snapshot $v\in\{\text{initial},\text{latest}\}$, the empirical distribution over semantic cluster labels. Section \ref{seq:equation} in Appendix provides the main equations of the methodology. Specifically, Normalized Shannon entropy $H_v(r)$ summarizes evenness (Eq.~\ref{eq:entropy}), and \emph{change} is $\Delta H(r)$ (Eq.~\ref{eq:deltaH}).
Distributional drift is measured with Jensen--Shannon divergence in bits between the aligned initial and latest distributions (Eq.~\ref{eq:jsd}).
Richness $K_v(r)$ is the count of distinct labels in snapshot $v$ with a presence threshold of at least two statements (Eq.~\ref{eq:richness}); the paired change is $\Delta K(r)$ (Eq.~\ref{eq:deltaK}).
To control for document length, we also report a rarefied $\Delta K$ by sampling an equal number of statements from both snapshots and averaging paired differences over repeated draws (Eq.~\ref{eq:rarefiedDK}).
Entropy and JSD are computed only for repositories with at least five labeled statements in each snapshot; richness uses the presence threshold described above.
All repository–level estimands are reported as equal-weight means across repositories with percentile confidence intervals obtained by a repository bootstrap (Eqs.~\ref{eq:bootReplicate}–\ref{eq:bootCI}; resampling repositories with replacement, $B{=}10{,}000$).
Unless noted otherwise, intervals are $95\%$ and units are bits for $H$ and JSD.

\section{Results}

We study within repository change by pairing the earliest recoverable governance snapshot with the latest and computing: (i) Shannon entropy for each version \(H_v(r)\) in Eq.~\eqref{eq:entropy} and the paired change \(\Delta H(r)\) in Eq.~\eqref{eq:deltaH}; (ii) distributional drift via Jensen–Shannon divergence in Eq.~\eqref{eq:jsd}; and (iii) breadth as the count of distinct constructs \(K_v(r)\) in Eq.~\eqref{eq:richness} and its paired change \(\Delta K(r)\) in Eq.~\eqref{eq:deltaK}, with the size controlled rarefied estimate \(\widetilde{\Delta K}(r)\) in Eq.~\eqref{eq:rarefiedDK}. Repository level summaries are aggregated with equal weight bootstrap confidence intervals using the resampling scheme in Eqs.~\eqref{eq:bootReplicate}–\eqref{eq:bootCI} \((B{=}10{,}000)\). Unless noted, units are bits for \(H\) and Jensen–Shannon divergence. The minimum per version screen of at least five labeled statements is applied as specified in Methods.

\textbf{Breadth.}
Projects define a wider array of who acts and what is governed over time. Roles and actions both show clear increases in the number of distinct constructs per repository, and those increases remain positive under the rarefied control that equalizes snapshot length \(\big(\widetilde{\Delta K}(r)\) in Eq.~\eqref{eq:rarefiedDK}\big). Deontic and object counts show no effect on average. Table~\ref{tab:count-results} reports the paired changes \(\Delta K\) with percentile confidence intervals that reflect between project variation.

\textbf{Concentration and drift.}
Attention across constructs becomes more evenly distributed for roles and actions. Mean \(\Delta H\) is positive for both features and the corresponding intervals exclude zero. Under the standard modal inventory deontic composition is broadly unchanged, while collapsing to enabling versus restricting yields a small decrease in evenness. Mean Jensen–Shannon divergence values indicate within project drift between initial and latest snapshots for all features, with larger drift for roles and actions than for deontics. Table~\ref{tab:entropy-results} reports \(\Delta H\) and Jensen–Shannon divergence.

%  ===== Table 2: Entropy (ΔH) and Drift (JSD)=====
\begin{table}[t]
\centering
\footnotesize
\setlength{\tabcolsep}{5pt}
\renewcommand{\arraystretch}{1.15}
\caption{\textbf{Attention to roles and actions becomes more even; deontic polarity is broadly stable.}
Entries report repository–paired changes in concentration (Shannon entropy \(\Delta H = H_{\text{latest}}-H_{\text{initial}}\), bits) (mean) and within–repository distributional drift (Jensen–Shannon divergence, bits). Rows show means across repositories; 95\% CIs are from equal–weight bootstrapping over repositories (\(B{=}10{,}000\)). Bold \(\Delta H\) intervals exclude 0. \(^{\dagger}\) Binary coding of deontics into enabling vs.\ restricting.}
\begin{tabular}{lcccccc}
\toprule
\textbf{Feature} & \textbf{$n$} & \textbf{Initial $H$} & \textbf{Latest $H$} & \textbf{$\Delta H$ [95\% CI]} & \textbf{JSD [95\% CI]} \\
\midrule
Roles   & 169 & 1.775 & 1.866 & \textbf{+0.092 [0.011, 0.173]} & 0.202 [0.172, 0.234] \\
Actions & 213 & 1.905 & 1.979 & \textbf{+0.074 [0.017, 0.134]} & 0.126 [0.107, 0.146] \\
Deontic & 144 & 1.057 & 1.052 & $-0.005$ [$-0.066$, 0.056]     & 0.062 [0.048, 0.079] \\
Deontic$^{\dagger}$ & 149 & 0.108 & 0.076 & \textbf{$-0.032$ [$-0.066$, $-0.001$]} & 0.009 [0.005, 0.014] \\
\bottomrule
\end{tabular}
\label{tab:entropy-results}
\end{table}

\textbf{Interpretation of intervals.}
The paired design means intervals summarize between project uncertainty, not within project sampling error. Results are robust to the minimum per version screen and rarefaction confirms that breadth findings are not an artifact of longer documents. Object results are underpowered in the paired subset and are reported for completeness.

Overall, communities diversify and rebalance the governance space of actors and activities while leaving prescriptive polarity comparatively stable. These patterns provide a human authored baseline against which AI assisted cohorts can be evaluated for concentration, drift, and breadth shifts \big(Tables~\ref{tab:count-results}–\ref{tab:entropy-results}\big).

% =========================
% Table 1: Count / Richness
% =========================
\begin{table}[t]
\centering
\footnotesize
\setlength{\tabcolsep}{6pt}
\renewcommand{\arraystretch}{1.15}
\caption{\textbf{Projects define more roles and govern more actions over time.}
Entries report the repository–paired change in the number of distinct constructs
\((\Delta K = K_{\text{latest}} - K_{\text{initial}})\)  (mean) with equal–weight bootstrap
95\% CIs over repositories (\(B{=}10{,}000\)). The rarefied estimate draws the same
number of statements from each snapshot (cap 100) before counting, to address
length differences. Units are counts of distinct clusters; bold intervals exclude 0.}
\label{tab:count-results}
\begin{tabular}{lccccc}
\toprule
\textbf{Feature} & \textbf{$n$} &
\textbf{Initial $K$} & \textbf{Latest $K$} &
\textbf{Mean $\Delta K$ [95\% CI]} & \textbf{Rarefied $\Delta K$ [95\% CI]} \\
\midrule
Roles    & 244 & 3.46 & 3.95 & \textbf{+0.484 [0.258, 0.713]} & \textbf{+0.224 [0.092, 0.352]} \\
Actions  & 266 & 3.86 & 4.46 & \textbf{+0.602 [0.417, 0.793]} & \textbf{+0.228 [0.134, 0.326]} \\
Deontics & 236 & 1.14 & 1.15 & +0.008 [$-0.038$, 0.055]        & $-0.024$ [$-0.062$, 0.012]      \\
Objects  & 97  & 1.27 & 1.33 & +0.062 [$-0.062$, 0.186]        & +0.075 [$-0.009$, 0.162]        \\
\bottomrule
\end{tabular}
\end{table}

\section{Interpretation}

Our paired design establishes a human authored baseline for how projects formalize participation and control over time. Two results are robust. Projects broaden and rebalance who acts and what is governed: both the count of distinct constructs and the evenness of attention rise for Roles and Actions (Tables~\ref{tab:count-results} and \ref{tab:entropy-results}). The force of rules is comparatively stable: under the standard Deontic inventory we see no effect on average, while an enabling versus restricting recode shows a small decrease in evenness. Objects are underpowered and we treat them as descriptive context.

These findings are consistent across complementary summaries. For Roles and Actions, entropy and richness move together, and rarefied $\Delta K$ confirms that gains are not explained by longer files. Jensen Shannon divergence indicates within repository drift as catalogs expand. The change distribution has many near zeros with a minority of large moves, which is consistent with punctuated edits rather than steady drift.

The measurements suggest priorities for AI assistance. Tools should emphasize specification over escalation: help communities name and rebalance Roles and Actions, surface uneven coverage, and propose governance ready statements that identify actor, deontic force, action, and object. Edits should be exposed as structured deltas that indicate which dimension changes. Given the small decrease in evenness under the enabling versus restricting recode, systems should require explicit acknowledgement before intensifying restricting language and make such shifts visible in review.

To evaluate AI assisted cohorts against the baseline, we recommend reporting paired changes in evenness and count for Roles and Actions (for example $\Delta H$ and $\Delta K$ with the rarefied variant), Deontic composition and polarity shares, authority concentration such as the share of approvals by role and the prevalence of single gate approvals, and participation outcomes such as time to first review and the distribution of review work. Where possible, control for repository size and activity.

Scope and external validity are limited. We analyze governance text rather than behavior, intervals reflect between project uncertainty via equal weight resampling at the repository level, and artifacts are predominantly in English with policies sometimes spread across files. The labeling, pairing, entropy and richness computation, Jensen Shannon divergence, and bootstrapping pipeline is artifact agnostic and can be applied to multi file policy inventories. Equal weighting avoids collapsing smaller projects into larger ones and aligns with participatory aims.

\section{Conclusion}

We provide a reproducible human baseline for how open-source projects formalize participation and control. Governance prose is parsed into tuples (actor, deontic force, action, object), clustered, and summarized with entropy \(H\) for evenness (Eq.~\ref{eq:entropy}), paired change \(\Delta H\) (Eq.~\ref{eq:deltaH}), Jensen–Shannon divergence for drift (Eq.~\ref{eq:jsd}), and richness \(K\) for the count of distinct constructs (Eq.~\ref{eq:richness}), with equal-weight repository bootstrapping for uncertainty (Eqs.~\ref{eq:bootReplicate}–\ref{eq:bootCI}). Empirically, projects define more \emph{roles} and govern more \emph{actions} over time (\(\Delta K>0\); rarefied estimates corroborate that gains are not length artifacts; Table~\ref{tab:count-results}); evenness also rises for roles and actions (\(\Delta H>0\)), while deontic composition is broadly stable, with a small decrease under the enabling/restricting recode (Table~\ref{tab:entropy-results}). Read as institutional signals, these patterns are consistent with maturation by accretion: catalogs of who acts and what is done broaden and rebalance, while prescriptive polarity changes slowly. Additionally, robustness checks across alternative minimum per-version thresholds showed that results remained consistent, indicating that observed patterns are not artifacts of threshold choice.

\noindent\textbf{Limitations.}
We analyze governance text rather than behavior; many consequential rules live outside \texttt{GOVERNANCE.md} (for example \texttt{CONTRIBUTING.md}, \texttt{CODEOWNERS}, CI settings, issue templates) or in informal channels, and the corpus is predominantly English open source, which limits generality.
Our paired design requires two recoverable snapshots per repository, so survivorship and timing effects can bias change estimates, and stabilization thresholds (at least five labeled statements per snapshot for evenness and drift, presence threshold $\tau=2$ for richness) trade variance for selection; Objects are comparatively sparse.
Natural language processing and representation choices, including segmentation, coreference, embedding, and clustering, can miss conditionality and shape absolute values of $K$, $H$, and Jensen Shannon divergence; although coder agreement for Action types was high, residual category bias is possible.
Inference is descriptive and comparative rather than causal: equal weight repository bootstrapping reflects between project variability without fully adjusting for confounders such as age or scale, and Jensen Shannon divergence is direction agnostic and can be affected by cluster relabeling across snapshots.

\textbf{Future Work.}
Researchers could extend this baseline to cohorts that adopt AI assisted governance and development tools, enabling before and after comparisons of concentration, breadth, and drift and linking governance change to outcomes such as review latency, newcomer acceptance, and workload distribution.  Researchers could also examine causal pathways by pairing textual change with event data from code review and release processes.

\section*{Acknowledgments}
This research was supported by the National Science Foundation under the Growing Convergence 
Research Scheme (Grant No.~2020751). Additional support was provided by NSF RCN Award 
No.~1917908, ``Coordinating and Advancing Analytical Approaches for Policy Design,'' and 
NSF Smart and Connected Communities Award No.~2421385, ``Designing Smart Environments 
to Augment Collective Learning and Creativity.'' We thank our collaborators and colleagues 
for their helpful discussions and feedback.

\bibliography{references}

\newpage
\appendix
\section*{Appendix}

\section{Methods}
Figure below presents an end-to-end pipeline that transforms raw governance documents into structured institutional insights. It begins with extracting initial and latest versions of GOVERNANCE.md files. Coreference resolution reduces pronoun ambiguity, enabling more accurate attribution of roles. Semantic Role Labeling maps sentences to predicate–argument structures, identifying the underlying grammar of governance actions. These structured tuples are then embedded and clustered using BERTopic to capture governance topologies. The resulting clusters are evaluated through per-project cluster counts, changes in the distribution of policy effort, and robustness checks. This modular pipeline supports scalable, interpretable analysis of institutional evolution in OSS projects.

% Place near where you first describe the pipeline
\begin{figure*}[!htb]
  \centering
  \includegraphics[width=\textwidth]{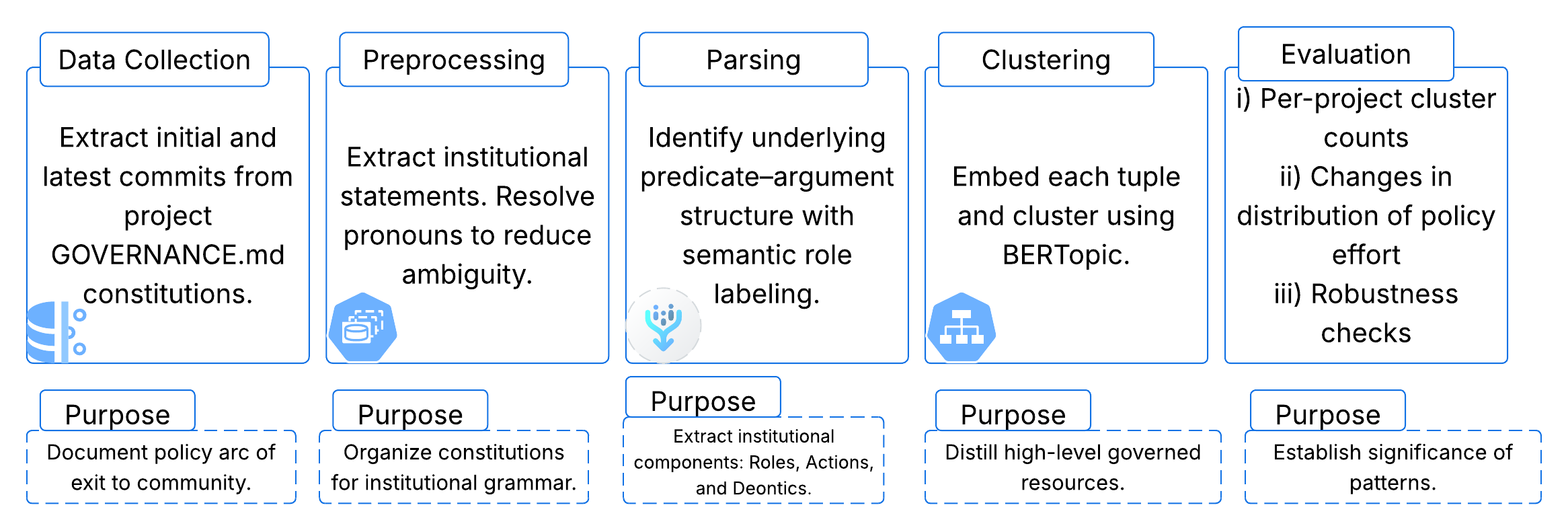}
  \vspace{-2mm}
  \caption{Processing pipeline from raw governance files to structured institutional statements and analysis. These steps support measurement of change in count and concentration from \textit{initial} to \textit{latest} versions of version-controlled \texttt{GOVERNANCE.md} project constitutions. This diagram shows how governance text is normalized, parsed into roles, actions, and deontics, and clustered into institutional constructs.}
  \label{fig:pipeline}
  \vspace{-3mm}
\end{figure*}

\label{seq:equation}

\noindent\fbox{\parbox{0.95\textwidth}{
\vspace{-0.3em}
\begin{gather}
H_v(r) \;=\; - \sum_{k} p_{r,v}(k)\,\log_2 p_{r,v}(k),
\quad v\in\{\text{initial},\text{latest}\}.\label{eq:entropy}\\[-0.3em]
\Delta H(r) \;=\; H_{\text{latest}}(r) \;-\; H_{\text{initial}}(r).\label{eq:deltaH}\\[-0.3em]
\mathrm{JSD}\!\left(p_{r,\text{initial}},p_{r,\text{latest}}\right)
\;=\; \tfrac{1}{2}\,\mathrm{KL}\!\left(p_{r,\text{initial}} \,\middle\|\, m_r\right)
\;+\; \tfrac{1}{2}\,\mathrm{KL}\!\left(p_{r,\text{latest}} \,\middle\|\, m_r\right),\nonumber\\
\quad m_r=\tfrac{1}{2}\!\left(p_{r,\text{initial}}+p_{r,\text{latest}}\right).\label{eq:jsd}\\[-0.3em]
K_v(r) \;=\; \sum_{k}\mathbf{1}\!\left\{\,c_{r,v}(k)\ge \tau\,\right\}, \qquad \tau=2.\label{eq:richness}\\[-0.3em]
\Delta K(r) \;=\; K_{\text{latest}}(r) \;-\; K_{\text{initial}}(r).\label{eq:deltaK}\\[-0.3em]
\widetilde{\Delta K}(r) \;=\; \frac{1}{R}\sum_{t=1}^{R}
\Big( K^{(t)}_{\text{latest}}(r;n_r) \;-\; K^{(t)}_{\text{initial}}(r;n_r) \Big),\nonumber\\
\quad n_r=\min\!\big\{N_{r,\text{initial}},\,N_{r,\text{latest}},\,100\big\}.\label{eq:rarefiedDK}\\[-0.3em]
\hat{\theta}^{*(b)} \;=\; \frac{1}{n}\sum_{r \in \mathcal{R}^{*(b)}} s(r),
\qquad b=1,\dots,B,\; B=10{,}000.\label{eq:bootReplicate}\\[-0.3em]
\mathrm{CI}_{1-\alpha} \;=\;
\Big[\, Q_{\alpha/2}\big(\{\hat{\theta}^{*(b)}\}\big),\;
Q_{1-\alpha/2}\big(\{\hat{\theta}^{*(b)}\}\big) \,\Big].\label{eq:bootCI}
\end{gather}
\vspace{-0.3em}
}}

\section{Computational Resource} 
Some experiments were run using Google Colab with freely available GPU resources. Additional analysis and processing were performed on a local machine equipped with four Quadro RTX 8000 GPUs (48GB VRAM each), CUDA version 12.4, and driver version 550.163.01. Resource usage remained moderate, and no large-scale distributed training was required.

\section{Additional Results}

\begin{figure*}[!htb]
  \centering

  \begin{subfigure}[t]{0.45\textwidth}
    \centering
    \includegraphics[width=0.9\textwidth, height=0.4\textwidth]{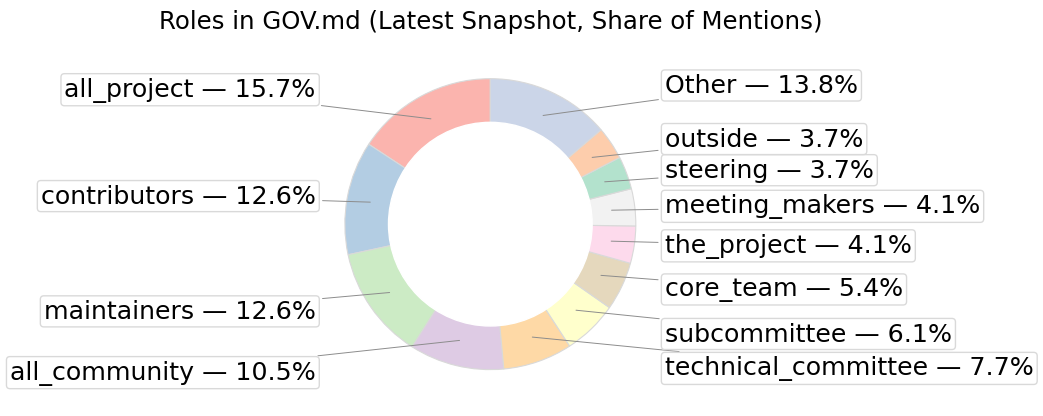}
    \subcaption{Roles}
    \label{fig:donut-roles}
  \end{subfigure}
  \hfill
  \begin{subfigure}[t]{0.45\textwidth}
    \centering
    \includegraphics[width=0.7\textwidth, height=0.4\textwidth]{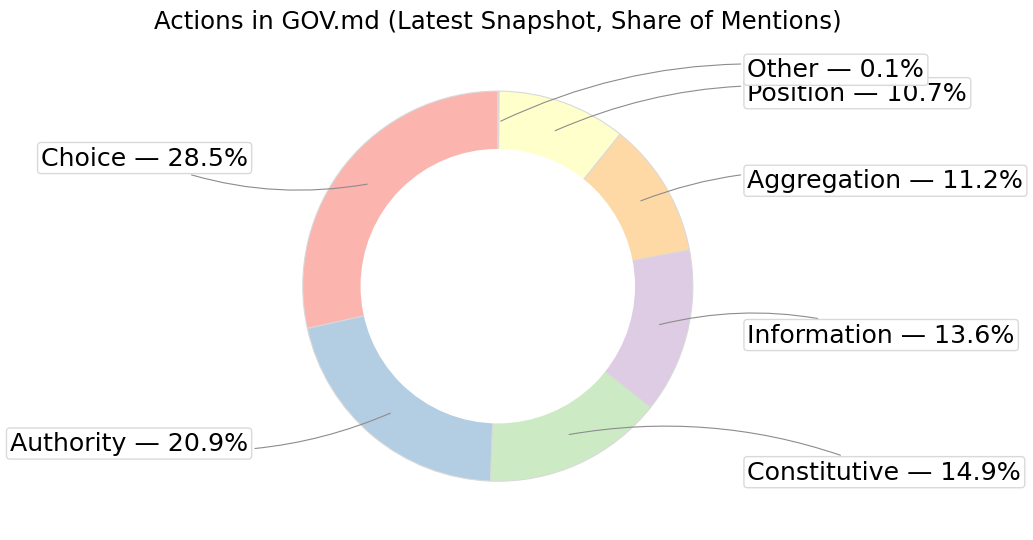}
    \subcaption{Actions}
    \label{fig:donut-actions}
  \end{subfigure}

  \vspace{1em}

  \begin{subfigure}[t]{0.45\textwidth}
    \centering
    \includegraphics[width=0.65\textwidth, height=0.4\textwidth]{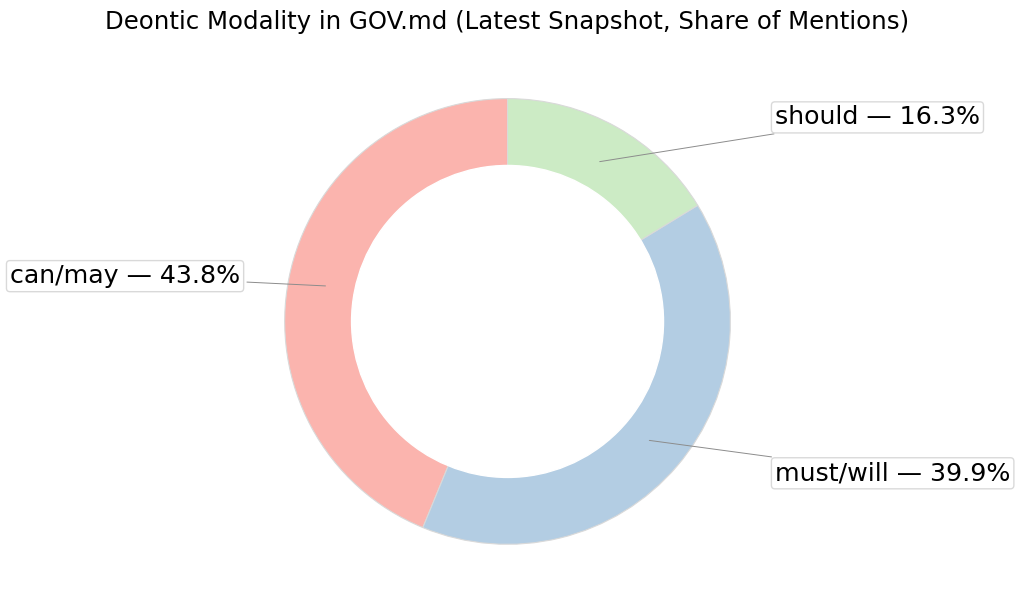}
    \subcaption{Deontics}
    \label{fig:donut-deontics}
  \end{subfigure}
  \hfill
  \begin{subfigure}[t]{0.45\textwidth}
    \centering
    \includegraphics[width=1\textwidth, height=0.4\textwidth]{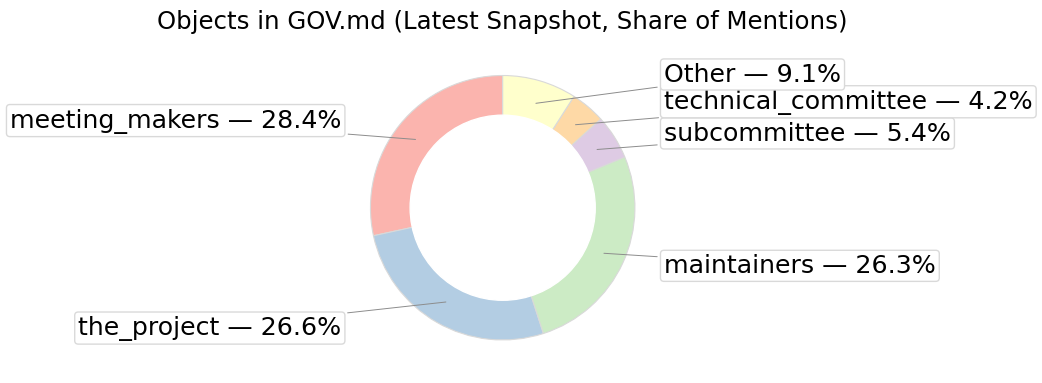}
    \subcaption{Objects}
    \label{fig:donut-objects}
  \end{subfigure}

  \caption{\textbf{Latest-snapshot composition of governance constructs.}
  Donuts show the relative share of clusters within each feature (Roles, Actions, Deontics, Objects).
  These panels are descriptive context; inference relies on paired change metrics reported in the main text.}
  \label{fig:donuts-latest-stacked}
\end{figure*}

% Appendix: Initial vs Latest violin plots (2×2)
\begin{figure*}[!htb]
  \centering

  % Row 1
  \begin{subfigure}[t]{0.48\linewidth}
    \centering
    \includegraphics[width=\linewidth]{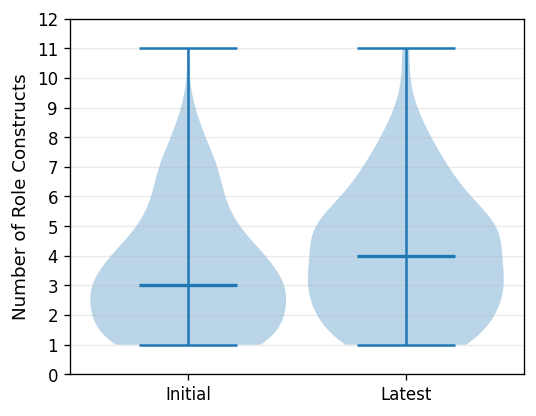}
    \subcaption{Roles}
    \label{fig:violins-roles}
  \end{subfigure}\hfill
  \begin{subfigure}[t]{0.48\linewidth}
    \centering
    \includegraphics[width=\linewidth]{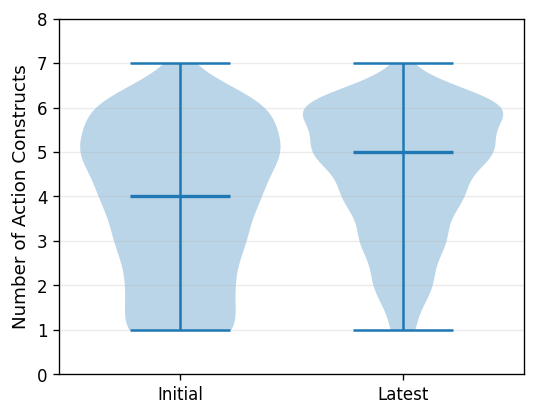}
    \subcaption{Actions}
    \label{fig:violins-actions}
  \end{subfigure}

  \vspace{0.8em}

  % Row 2
  \begin{subfigure}[t]{0.48\linewidth}
    \centering
    \includegraphics[width=\linewidth]{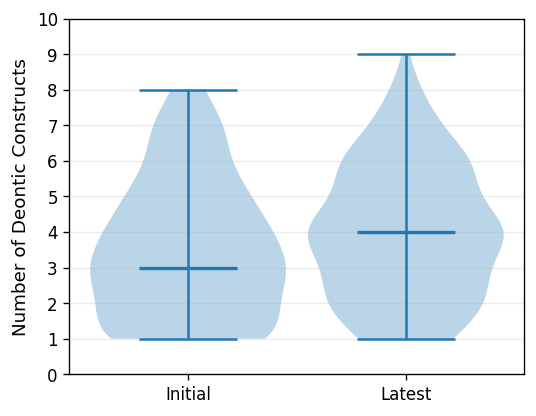}
    \subcaption{Deontics}
    \label{fig:violins-deontics}
  \end{subfigure}\hfill
  \begin{subfigure}[t]{0.48\linewidth}
    \centering
    \includegraphics[width=\linewidth]{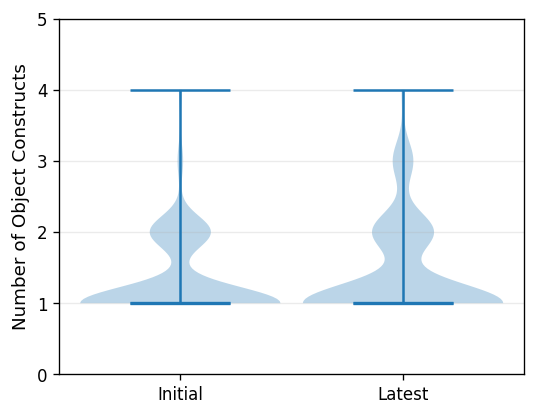}
    \subcaption{Objects}
    \label{fig:violins-objects}
  \end{subfigure}

  \caption{Distribution of per-repository \textbf{distinct construct counts} at the initial and latest snapshots for each institutional feature. Violins show density; the horizontal line is the median. Panels provide descriptive context; paired bootstrap estimates are reported in Table~\ref{tab:count-results}.}
  \label{fig:violins-initial-latest-2x2}
\end{figure*}

\newpage
\section*{NeurIPS Paper Checklist}

%%% BEGIN INSTRUCTIONS %%%
The checklist is designed to encourage best practices for responsible machine learning research, addressing issues of reproducibility, transparency, research ethics, and societal impact. Do not remove the checklist: {\bf The papers not including the checklist will be desk rejected.} The checklist should follow the references and follow the (optional) supplemental material.  The checklist does NOT count towards the page
limit. 

Please read the checklist guidelines carefully for information on how to answer these questions. For each question in the checklist:
\begin{itemize}
    \item You should answer \answerYes{}, \answerNo{}, or \answerNA{}.
    \item \answerNA{} means either that the question is Not Applicable for that particular paper or the relevant information is Not Available.
    \item Please provide a short (1–2 sentence) justification right after your answer (even for NA). 
   % \item {\bf The papers not including the checklist will be desk rejected.}
\end{itemize}

{\bf The checklist answers are an integral part of your paper submission.} They are visible to the reviewers, area chairs, senior area chairs, and ethics reviewers. You will be asked to also include it (after eventual revisions) with the final version of your paper, and its final version will be published with the paper.

The reviewers of your paper will be asked to use the checklist as one of the factors in their evaluation. While "\answerYes{}" is generally preferable to "\answerNo{}", it is perfectly acceptable to answer "\answerNo{}" provided a proper justification is given (e.g., "error bars are not reported because it would be too computationally expensive" or "we were unable to find the license for the dataset we used"). In general, answering "\answerNo{}" or "\answerNA{}" is not grounds for rejection. While the questions are phrased in a binary way, we acknowledge that the true answer is often more nuanced, so please just use your best judgment and write a justification to elaborate. All supporting evidence can appear either in the main paper or the supplemental material, provided in appendix. If you answer \answerYes{} to a question, in the justification please point to the section(s) where related material for the question can be found.

IMPORTANT, please:
\begin{itemize}
    \item {\bf Delete this instruction block, but keep the section heading ``NeurIPS Paper Checklist"},
    \item  {\bf Keep the checklist subsection headings, questions/answers and guidelines below.}
    \item {\bf Do not modify the questions and only use the provided macros for your answers}.
\end{itemize}

%%% END INSTRUCTIONS %%%

\begin{enumerate}

\item {\bf Claims}
    \item[] Question: Do the main claims made in the abstract and introduction accurately reflect the paper's contributions and scope?
    \item[] Answer: \answerYes{} % Replace by \answerYes{}, \answerNo{}, or \answerNA{}.
    \item[] Justification: Sections 1-5
    \item[] Guidelines:
    \begin{itemize}
        \item The answer NA means that the abstract and introduction do not include the claims made in the paper.
        \item The abstract and/or introduction should clearly state the claims made, including the contributions made in the paper and important assumptions and limitations. A No or NA answer to this question will not be perceived well by the reviewers. 
        \item The claims made should match theoretical and experimental results, and reflect how much the results can be expected to generalize to other settings. 
        \item It is fine to include aspirational goals as motivation as long as it is clear that these goals are not attained by the paper. 
    \end{itemize}

\item {\bf Limitations}
    \item[] Question: Does the paper discuss the limitations of the work performed by the authors?
    \item[] Answer: \answerYes{} % Replace by \answerYes{}, \answerNo{}, or \answerNA{}.
    \item[] Justification: Section 5
    \item[] Guidelines:
    \begin{itemize}
        \item The answer NA means that the paper has no limitation while the answer No means that the paper has limitations, but those are not discussed in the paper. 
        \item The authors are encouraged to create a separate "Limitations" section in their paper.
        \item The paper should point out any strong assumptions and how robust the results are to violations of these assumptions (e.g., independence assumptions, noiseless settings, model well-specification, asymptotic approximations only holding locally). The authors should reflect on how these assumptions might be violated in practice and what the implications would be.
        \item The authors should reflect on the scope of the claims made, e.g., if the approach was only tested on a few datasets or with a few runs. In general, empirical results often depend on implicit assumptions, which should be articulated.
        \item The authors should reflect on the factors that influence the performance of the approach. For example, a facial recognition algorithm may perform poorly when image resolution is low or images are taken in low lighting. Or a speech-to-text system might not be used reliably to provide closed captions for online lectures because it fails to handle technical jargon.
        \item The authors should discuss the computational efficiency of the proposed algorithms and how they scale with dataset size.
        \item If applicable, the authors should discuss possible limitations of their approach to address problems of privacy and fairness.
        \item While the authors might fear that complete honesty about limitations might be used by reviewers as grounds for rejection, a worse outcome might be that reviewers discover limitations that aren't acknowledged in the paper. The authors should use their best judgment and recognize that individual actions in favor of transparency play an important role in developing norms that preserve the integrity of the community. Reviewers will be specifically instructed to not penalize honesty concerning limitations.
    \end{itemize}

\item {\bf Theory assumptions and proofs}
    \item[] Question: For each theoretical result, does the paper provide the full set of assumptions and a complete (and correct) proof?
    \item[] Answer: \answerNA{} % Replace by \answerYes{}, \answerNo{}, or \answerNA{}.
    \item[] Justification: \answerNA{}
    \item[] Guidelines:
    \begin{itemize}
        \item The answer NA means that the paper does not include theoretical results. 
        \item All the theorems, formulas, and proofs in the paper should be numbered and cross-referenced.
        \item All assumptions should be clearly stated or referenced in the statement of any theorems.
        \item The proofs can either appear in the main paper or the supplemental material, but if they appear in the supplemental material, the authors are encouraged to provide a short proof sketch to provide intuition. 
        \item Inversely, any informal proof provided in the core of the paper should be complemented by formal proofs provided in appendix or supplemental material.
        \item Theorems and Lemmas that the proof relies upon should be properly referenced. 
    \end{itemize}

    \item {\bf Experimental result reproducibility}
    \item[] Question: Does the paper fully disclose all the information needed to reproduce the main experimental results of the paper to the extent that it affects the main claims and/or conclusions of the paper (regardless of whether the code and data are provided or not)?
    \item[] Answer: \answerYes{} % Replace by \answerYes{}, \answerNo{}, or \answerNA{}.
    \item[] Justification: Sections 2 and 3
    \item[] Guidelines:
    \begin{itemize}
        \item The answer NA means that the paper does not include experiments.
        \item If the paper includes experiments, a No answer to this question will not be perceived well by the reviewers: Making the paper reproducible is important, regardless of whether the code and data are provided or not.
        \item If the contribution is a dataset and/or model, the authors should describe the steps taken to make their results reproducible or verifiable. 
        \item Depending on the contribution, reproducibility can be accomplished in various ways. For example, if the contribution is a novel architecture, describing the architecture fully might suffice, or if the contribution is a specific model and empirical evaluation, it may be necessary to either make it possible for others to replicate the model with the same dataset, or provide access to the model. In general. releasing code and data is often one good way to accomplish this, but reproducibility can also be provided via detailed instructions for how to replicate the results, access to a hosted model (e.g., in the case of a large language model), releasing of a model checkpoint, or other means that are appropriate to the research performed.
        \item While NeurIPS does not require releasing code, the conference does require all submissions to provide some reasonable avenue for reproducibility, which may depend on the nature of the contribution. For example
        \begin{enumerate}
            \item If the contribution is primarily a new algorithm, the paper should make it clear how to reproduce that algorithm.
            \item If the contribution is primarily a new model architecture, the paper should describe the architecture clearly and fully.
            \item If the contribution is a new model (e.g., a large language model), then there should either be a way to access this model for reproducing the results or a way to reproduce the model (e.g., with an open-source dataset or instructions for how to construct the dataset).
            \item We recognize that reproducibility may be tricky in some cases, in which case authors are welcome to describe the particular way they provide for reproducibility. In the case of closed-source models, it may be that access to the model is limited in some way (e.g., to registered users), but it should be possible for other researchers to have some path to reproducing or verifying the results.
        \end{enumerate}
    \end{itemize}

\item {\bf Open access to data and code}
    \item[] Question: Does the paper provide open access to the data and code, with sufficient instructions to faithfully reproduce the main experimental results, as described in supplemental material?
    \item[] Answer: \answerYes{} % Replace by \answerYes{}, \answerNo{}, or \answerNA{}.
    \item[] Justification: Data is public.
    \item[] Guidelines:
    \begin{itemize}
        \item The answer NA means that paper does not include experiments requiring code.
        \item Please see the NeurIPS code and data submission guidelines (\url{https://nips.cc/public/guides/CodeSubmissionPolicy}) for more details.
        \item While we encourage the release of code and data, we understand that this might not be possible, so “No” is an acceptable answer. Papers cannot be rejected simply for not including code, unless this is central to the contribution (e.g., for a new open-source benchmark).
        \item The instructions should contain the exact command and environment needed to run to reproduce the results. See the NeurIPS code and data submission guidelines (\url{https://nips.cc/public/guides/CodeSubmissionPolicy}) for more details.
        \item The authors should provide instructions on data access and preparation, including how to access the raw data, preprocessed data, intermediate data, and generated data, etc.
        \item The authors should provide scripts to reproduce all experimental results for the new proposed method and baselines. If only a subset of experiments are reproducible, they should state which ones are omitted from the script and why.
        \item At submission time, to preserve anonymity, the authors should release anonymized versions (if applicable).
        \item Providing as much information as possible in supplemental material (appended to the paper) is recommended, but including URLs to data and code is permitted.
    \end{itemize}

\item {\bf Experimental setting/details}
    \item[] Question: Does the paper specify all the training and test details (e.g., data splits, hyperparameters, how they were chosen, type of optimizer, etc.) necessary to understand the results?
    \item[] Answer: \answerYes{} % Replace by \answerYes{}, \answerNo{}, or \answerNA{}.
    \item[] Justification: Section 2
    \item[] Guidelines:
    \begin{itemize}
        \item The answer NA means that the paper does not include experiments.
        \item The experimental setting should be presented in the core of the paper to a level of detail that is necessary to appreciate the results and make sense of them.
        \item The full details can be provided either with the code, in appendix, or as supplemental material.
    \end{itemize}

\item {\bf Experiment statistical significance}
    \item[] Question: Does the paper report error bars suitably and correctly defined or other appropriate information about the statistical significance of the experiments?
    \item[] Answer: \answerNA{} % Replace by \answerYes{}, \answerNo{}, or \answerNA{}.
    \item[] Justification: \answerNA{}
    \item[] Guidelines:
    \begin{itemize}
        \item The answer NA means that the paper does not include experiments.
        \item The authors should answer "Yes" if the results are accompanied by error bars, confidence intervals, or statistical significance tests, at least for the experiments that support the main claims of the paper.
        \item The factors of variability that the error bars are capturing should be clearly stated (for example, train/test split, initialization, random drawing of some parameter, or overall run with given experimental conditions).
        \item The method for calculating the error bars should be explained (closed form formula, call to a library function, bootstrap, etc.)
        \item The assumptions made should be given (e.g., Normally distributed errors).
        \item It should be clear whether the error bar is the standard deviation or the standard error of the mean.
        \item It is OK to report 1-sigma error bars, but one should state it. The authors should preferably report a 2-sigma error bar than state that they have a 96\% CI, if the hypothesis of Normality of errors is not verified.
        \item For asymmetric distributions, the authors should be careful not to show in tables or figures symmetric error bars that would yield results that are out of range (e.g. negative error rates).
        \item If error bars are reported in tables or plots, The authors should explain in the text how they were calculated and reference the corresponding figures or tables in the text.
    \end{itemize}

\item {\bf Experiments compute resources}
    \item[] Question: For each experiment, does the paper provide sufficient information on the computer resources (type of compute workers, memory, time of execution) needed to reproduce the experiments?
    \item[] Answer: \answerYes{} % Replace by \answerYes{}, \answerNo{}, or \answerNA{}.
    \item[] Justification: Appendix
    \item[] Guidelines:
    \begin{itemize}
        \item The answer NA means that the paper does not include experiments.
        \item The paper should indicate the type of compute workers CPU or GPU, internal cluster, or cloud provider, including relevant memory and storage.
        \item The paper should provide the amount of compute required for each of the individual experimental runs as well as estimate the total compute. 
        \item The paper should disclose whether the full research project required more compute than the experiments reported in the paper (e.g., preliminary or failed experiments that didn't make it into the paper). 
    \end{itemize}
    
\item {\bf Code of ethics}
    \item[] Question: Does the research conducted in the paper conform, in every respect, with the NeurIPS Code of Ethics \url{https://neurips.cc/public/EthicsGuidelines}?
    \item[] Answer: \answerYes{} % Replace by \answerYes{}, \answerNo{}, or \answerNA{}.
    \item[] Justification: \answerYes{}
    \item[] Guidelines:
    \begin{itemize}
        \item The answer NA means that the authors have not reviewed the NeurIPS Code of Ethics.
        \item If the authors answer No, they should explain the special circumstances that require a deviation from the Code of Ethics.
        \item The authors should make sure to preserve anonymity (e.g., if there is a special consideration due to laws or regulations in their jurisdiction).
    \end{itemize}

\item {\bf Broader impacts}
    \item[] Question: Does the paper discuss both potential positive societal impacts and negative societal impacts of the work performed?
    \item[] Answer: \answerYes{} % Replace by \answerYes{}, \answerNo{}, or \answerNA{}.
    \item[] Justification: Sections 4 and 5
    \item[] Guidelines:
    \begin{itemize}
        \item The answer NA means that there is no societal impact of the work performed.
        \item If the authors answer NA or No, they should explain why their work has no societal impact or why the paper does not address societal impact.
        \item Examples of negative societal impacts include potential malicious or unintended uses (e.g., disinformation, generating fake profiles, surveillance), fairness considerations (e.g., deployment of technologies that could make decisions that unfairly impact specific groups), privacy considerations, and security considerations.
        \item The conference expects that many papers will be foundational research and not tied to particular applications, let alone deployments. However, if there is a direct path to any negative applications, the authors should point it out. For example, it is legitimate to point out that an improvement in the quality of generative models could be used to generate deepfakes for disinformation. On the other hand, it is not needed to point out that a generic algorithm for optimizing neural networks could enable people to train models that generate Deepfakes faster.
        \item The authors should consider possible harms that could arise when the technology is being used as intended and functioning correctly, harms that could arise when the technology is being used as intended but gives incorrect results, and harms following from (intentional or unintentional) misuse of the technology.
        \item If there are negative societal impacts, the authors could also discuss possible mitigation strategies (e.g., gated release of models, providing defenses in addition to attacks, mechanisms for monitoring misuse, mechanisms to monitor how a system learns from feedback over time, improving the efficiency and accessibility of ML).
    \end{itemize}
    
\item {\bf Safeguards}
    \item[] Question: Does the paper describe safeguards that have been put in place for responsible release of data or models that have a high risk for misuse (e.g., pretrained language models, image generators, or scraped datasets)?
    \item[] Answer: \answerNA{} % Replace by \answerYes{}, \answerNo{}, or \answerNA{}.
    \item[] Justification: \answerNA{}
    \item[] Guidelines:
    \begin{itemize}
        \item The answer NA means that the paper poses no such risks.
        \item Released models that have a high risk for misuse or dual-use should be released with necessary safeguards to allow for controlled use of the model, for example by requiring that users adhere to usage guidelines or restrictions to access the model or implementing safety filters. 
        \item Datasets that have been scraped from the Internet could pose safety risks. The authors should describe how they avoided releasing unsafe images.
        \item We recognize that providing effective safeguards is challenging, and many papers do not require this, but we encourage authors to take this into account and make a best faith effort.
    \end{itemize}

\item {\bf Licenses for existing assets}
    \item[] Question: Are the creators or original owners of assets (e.g., code, data, models), used in the paper, properly credited and are the license and terms of use explicitly mentioned and properly respected?
    \item[] Answer: \answerYes{} % Replace by \answerYes{}, \answerNo{}, or \answerNA{}.
    \item[] Justification: Sections 1 and 2
    \item[] Guidelines:
    \begin{itemize}
        \item The answer NA means that the paper does not use existing assets.
        \item The authors should cite the original paper that produced the code package or dataset.
        \item The authors should state which version of the asset is used and, if possible, include a URL.
        \item The name of the license (e.g., CC-BY 4.0) should be included for each asset.
        \item For scraped data from a particular source (e.g., website), the copyright and terms of service of that source should be provided.
        \item If assets are released, the license, copyright information, and terms of use in the package should be provided. For popular datasets, \url{paperswithcode.com/datasets} has curated licenses for some datasets. Their licensing guide can help determine the license of a dataset.
        \item For existing datasets that are re-packaged, both the original license and the license of the derived asset (if it has changed) should be provided.
        \item If this information is not available online, the authors are encouraged to reach out to the asset's creators.
    \end{itemize}

\item {\bf New assets}
    \item[] Question: Are new assets introduced in the paper well documented and is the documentation provided alongside the assets?
    \item[] Answer: \answerNA{} % Replace by \answerYes{}, \answerNo{}, or \answerNA{}.
    \item[] Justification: \answerNA{}
    \item[] Guidelines:
    \begin{itemize}
        \item The answer NA means that the paper does not release new assets.
        \item Researchers should communicate the details of the dataset/code/model as part of their submissions via structured templates. This includes details about training, license, limitations, etc. 
        \item The paper should discuss whether and how consent was obtained from people whose asset is used.
        \item At submission time, remember to anonymize your assets (if applicable). You can either create an anonymized URL or include an anonymized zip file.
    \end{itemize}

\item {\bf Crowdsourcing and research with human subjects}
    \item[] Question: For crowdsourcing experiments and research with human subjects, does the paper include the full text of instructions given to participants and screenshots, if applicable, as well as details about compensation (if any)? 
    \item[] Answer: \answerNA{} % Replace by \answerYes{}, \answerNo{}, or \answerNA{}.
    \item[] Justification: \answerNA{}
    \item[] Guidelines:
    \begin{itemize}
        \item The answer NA means that the paper does not involve crowdsourcing nor research with human subjects.
        \item Including this information in the supplemental material is fine, but if the main contribution of the paper involves human subjects, then as much detail as possible should be included in the main paper. 
        \item According to the NeurIPS Code of Ethics, workers involved in data collection, curation, or other labor should be paid at least the minimum wage in the country of the data collector. 
    \end{itemize}

\item {\bf Institutional review board (IRB) approvals or equivalent for research with human subjects}
    \item[] Question: Does the paper describe potential risks incurred by study participants, whether such risks were disclosed to the subjects, and whether Institutional Review Board (IRB) approvals (or an equivalent approval/review based on the requirements of your country or institution) were obtained?
    \item[] Answer: \answerNA{} % Replace by \answerYes{}, \answerNo{}, or \answerNA{}.
    \item[] Justification: \answerNA{}
    \item[] Guidelines:
    \begin{itemize}
        \item The answer NA means that the paper does not involve crowdsourcing nor research with human subjects.
        \item Depending on the country in which research is conducted, IRB approval (or equivalent) may be required for any human subjects research. If you obtained IRB approval, you should clearly state this in the paper. 
        \item We recognize that the procedures for this may vary significantly between institutions and locations, and we expect authors to adhere to the NeurIPS Code of Ethics and the guidelines for their institution. 
        \item For initial submissions, do not include any information that would break anonymity (if applicable), such as the institution conducting the review.
    \end{itemize}

\item {\bf Declaration of LLM usage}
    \item[] Question: Does the paper describe the usage of LLMs if it is an important, original, or non-standard component of the core methods in this research? Note that if the LLM is used only for writing, editing, or formatting purposes and does not impact the core methodology, scientific rigorousness, or originality of the research, declaration is not required.
    %this research? 
    \item[] Answer: \answerNA{} % Replace by \answerYes{}, \answerNo{}, or \answerNA{}.
    \item[] Justification: \answerNA{}
    \item[] Guidelines:
    \begin{itemize}
        \item The answer NA means that the core method development in this research does not involve LLMs as any important, original, or non-standard components.
        \item Please refer to our LLM policy (\url{https://neurips.cc/Conferences/2025/LLM}) for what should or should not be described.
    \end{itemize}

\end{enumerate}

\end{document}